\documentclass{article}
\usepackage{ijcai26}
\usepackage{times}
\usepackage{soul}
\usepackage{url}
\usepackage[hidelinks]{hyperref}
\usepackage[utf8]{inputenc}
\usepackage[small]{caption}
\usepackage{graphicx}
\usepackage{amsmath}
\usepackage{amsthm}
\usepackage{booktabs}
\usepackage{algorithm}
\usepackage{algorithmic}
\usepackage[switch]{lineno}
\usepackage{amssymb}
\usepackage{bm}
\usepackage{multirow, multicol}
\usepackage{latexsym}
\usepackage{microtype}
\usepackage{inconsolata}
\usepackage{array}
\usepackage{tabularx}
\PassOptionsToPackage{table}{xcolor}
\usepackage{tikz}
\usetikzlibrary{positioning,fit,arrows.meta,calc}
\theoremstyle{plain}

\definecolor{AlgoBG}{RGB}{235,243,255}   
\definecolor{ArchBG}{RGB}{235,250,240}   
\definecolor{ScaleBG}{RGB}{255,244,232}  

\title{Accelerating Masked Diffusion Large Language Models: \\ A Survey of Efficient Inference Techniques}

\author{
Daehoon Gwak$^1$, Minhyung Lee$^2$, Junwoo Park$^1$, Jaegul Choo$^1$\\
\affiliations
$^1$KAIST AI\\
$^2$Yonsei University\\
\emails
daehoon.gwak@kaist.ac.kr, jchoo@kaist.ac.kr
}

\begin{document}

\maketitle

\begin{abstract}
Diffusion large language models (dLLMs) offer a theoretical advantage in parallel generation over standard autoregressive models. However, parallel generation alone does not guarantee practical speedups. Realizing this efficiency requires specialized inference mechanisms, such as diffusion-aware caching and reuse. Consequently, as inference efficiency becomes a prerequisite for practical deployment, recent research has actively explored acceleration techniques across algorithms, architectures, and systems. However, rigorous comparisons remain difficult, as end-to-end latency stems from intricate trade-offs between algorithmic, architectural, and system-level factors that are often conflated in existing benchmarks. In this survey, we introduce a unified latency decomposition framework for dLLMs to disentangle these factors and analyze their impact on inference speed in real deployments. Guided by this framework, we categorize acceleration techniques along three axes covering algorithmic innovations, architectural and system optimizations, and inference-time scaling. Finally, we provide guidelines for reproducible benchmarking and highlight open challenges for realizing the full potential of parallel generation.
\end{abstract}

\section{Introduction}
Autoregressive (AR) models remain the dominant paradigm for large language models. However, their inference cost scales linearly with the generated length. Specifically, producing $L$ tokens requires at least $L$ dependent decoding iterations.
This sequential dependency is a fundamental barrier for latency-sensitive applications and high-throughput serving, motivating extensive research on parallel generation.
Recently, Diffusion-based large language models (dLLMs) \cite{Nie2025Large,Ye2025Dream,Labs2025Mercury} have emerged as a promising alternative. They replace strict left-to-right decoding with iterative refinement, enabling parallel updates across multiple token positions \cite{Austin2021Structured,Lou2024Discrete,Sahoo2024Simple}.

However, parallel generation alone does not guarantee practical speedups. Realizing this efficiency requires specialized inference mechanisms, such as diffusion-aware caching and reuse, which introduce their own complexities. Consequently, as inference efficiency becomes a prerequisite for practical deployment, recent research has actively explored acceleration techniques across algorithms, architectures, and systems. However, rigorous comparisons remain difficult, as end-to-end latency stems from intricate trade-offs between these factors. For instance, an algorithmic method might reduce refinement steps but increase per-step policy overhead, while a system optimization might improve throughput at the cost of memory pressure. These algorithmic, architectural, and system-level impacts are often conflated in existing benchmarks, making it challenging to isolate the true sources of performance gains in real-world deployments.

To address these challenges, this survey focuses specifically on inference-time efficiency in diffusion-based language generation.
While recent surveys provide broad overviews of diffusion language modeling~\cite{Li2025Survey} or parallel generation mechanisms~\cite{Zhang2025Survey}, our work synthesizes the literature through a distinct, deployment-oriented efficiency lens. Specifically, we aim to clarify: (i) the determinants of end-to-end latency, throughput, and memory in realistic workloads; (ii) the inherent trade-offs between quality, speed, and memory across different acceleration techniques; and (iii) the best practices for reproducible benchmarking to ensure meaningful comparisons.

\begin{table*}[t]
\centering
\small
\setlength{\tabcolsep}{4pt}
\renewcommand{\arraystretch}{1.05}
\begin{tabularx}{\textwidth}{p{3.05cm} p{3.7cm} X >{\centering\arraybackslash}p{3.0cm}}
\toprule
\textbf{Family} &
\textbf{Primary levers (Eq.~\ref{eq:latency_decomposition})} &
\textbf{Key idea} &
\textbf{Section} \\ 
\midrule

\rowcolor{AlgoBG}
\multicolumn{4}{l}{\textbf{Algorithmic efficiency}}\\
\addlinespace[2pt]
Schedules \& policies &
$T\downarrow$, $C_{\mathrm{policy}}\uparrow$ &
Fewer steps via confidence/dilated schedules and selective updates; overhead depends on implementation. &
Section~\ref{sec:algo_schedules} \\

Decoding algorithm &
$T\downarrow$ or $G_t\downarrow$, $C_{\mathrm{policy}}\uparrow$ &
Larger parallel progress or draft-and-verify; gains hinge on acceptance and orchestration cost. &
Section~\ref{sec:algo_speculative} \\

Distillation / consistency &
$T\downarrow$ &
Few-step samplers shift cost to training; quality/calibration can become sensitive. &
Section~\ref{sec:algo_distill} \\

\addlinespace[2pt]
\rowcolor{ArchBG}
\multicolumn{4}{l}{\textbf{Architectural \& systems efficiency}}\\
\addlinespace[2pt]
Architecture \& numerics &
$C_{\mathrm{fwd}}\downarrow$, Mem$\downarrow$ &
Reduce per-forward cost via sparse/structured decoding or quantization; depends on kernel maturity. &
Section~\ref{sec:arch_sparsity} \\

Caching \& reuse &
$C_{\mathrm{fwd}}\downarrow$, $C_{\mathrm{sys}}\downarrow$, Mem$\uparrow$ &
Reuse KV/activations across refinement steps; explicit memory--speed trade-off. &
Section~\ref{sec:arch_cache} \\

System-level serving &
$C_{\mathrm{sys}}\downarrow$ &
Kernel/orchestration/microbatching improvements matter most for batch-1 and deployment regimes. &
Section~\ref{sec:arch_system} \\

\rowcolor{ScaleBG}
\multicolumn{4}{l}{\textbf{Inference-time scaling}}\\
\addlinespace[2pt]
Guidance \& search &
$G_t\uparrow$, $C_{\mathrm{policy}}\uparrow$, Mem$\uparrow$ &
Multi-pass decoding improves quality/constraints but increases total evaluations and memory pressure. &
Section~\ref{sec:scaling} \\

\bottomrule
\end{tabularx}
\caption{Compact taxonomy of inference-efficiency techniques for diffusion LLMs, organized by the latency decomposition in Eq.~\ref{eq:latency_decomposition}. Each family is summarized by the primary terms it tends to affect (arrows indicate the typical direction at a matched quality target), and the last column points to the section where we discuss representative methods, trade-offs, and evaluation considerations in detail.}
\label{tab:efficiency_levers}
\end{table*}

This survey makes three primary contributions:
\begin{itemize}
    \item We introduce a unified latency decomposition framework tailored to dLLMs, providing a rigorous basis for analyzing inference efficiency beyond nominal step counts.
    \item We propose a structured taxonomy that categorizes acceleration techniques into algorithmic, architectural, and system-level optimizations, explicitly mapping them to the terms in our efficiency framework.
    \item We establish best practices for reproducible benchmarking, covering latency, throughput, and memory measurement, and identify key open challenges at the intersection of algorithms and systems.
\end{itemize}

The remainder of this survey is organized as follows. We begin by reviewing the fundamentals of discrete diffusion decoding (Section~\ref{sec:background}) and presenting our latency decomposition framework (Section~\ref{sec:latency}). We then survey acceleration methods, grouped into algorithmic innovations (Section~\ref{sec:algorithmic}), architectural and system optimizations (Section~\ref{sec:archsys}), and inference-time scaling strategies (Section~\ref{sec:scaling}). Finally, we provide a practitioner's guide for interpreting efficiency claims and mapping objectives to the levers in Eq.~\eqref{eq:latency_decomposition} (Section~\ref{sec:guide_and_challenges}), and discuss future research directions (Section~\ref{sec:conclusion}).

\section{Background: Discrete Diffusion Decoding for Language}
\label{sec:background}

This section reviews essential mechanics and notation of discrete diffusion models needed for our \emph{inference-time efficiency} discussion.
We focus on the common \emph{masked} discrete diffusion setting, since it induces an explicit refinement loop with repeated Transformer evaluations which is the primary bottleneck addressed by recent acceleration methods.

\paragraph{Setup and notation.}
Let $\mathcal{V}$ be a vocabulary and let $\mathbf{x}=(x_1,\dots,x_L)\in\mathcal{V}^L$ denote a length-$L$ generated span, typically conditioned on a prompt or context $\mathbf{c}$.
When discussing inference cost, we distinguish the generated length $L$ from the total length processed by the model per evaluation, $L_{\mathrm{in}}$ (e.g., prompt length plus the generated span).
Discrete diffusion introduces a sequence of states $\mathbf{x}_T,\mathbf{x}_{T-1},\ldots,\mathbf{x}_0$, where larger $t$ indicates heavier corruption; $\mathbf{x}_0$ is the final output and $\mathbf{x}_T$ is commonly an all-\texttt{[MASK]} template.
We write $\mathcal{M}_t\subseteq\{1,\dots,L\}$ for the masked positions at step $t$.

A standard forward corruption in this setting is \emph{absorbing} masking: as $t$ increases, a subset of positions is replaced by \texttt{[MASK]}, yielding a Markov chain over discrete states \cite{Austin2021Structured}.
A corruption schedule controls how the expected mask ratio evolves with $t$, which determines how much information remains visible at intermediate states.
Given a partially corrupted sequence $\mathbf{x}_t$ and condition $\mathbf{c}$, the denoising model predicts token distributions, often for all positions even if the loss is applied only on masked positions \cite{Sahoo2024Simple}.
For our purposes, the key operational point is that inference repeatedly evaluates the model under changing corruption patterns, and each evaluation typically processes a sequence of length $L_{\mathrm{in}}$.

\paragraph{Reverse-time decoding and update sparsity.}
At inference time, decoding starts from $\mathbf{x}_T$ (often all \texttt{[MASK]}) and iteratively refines toward $\mathbf{x}_0$.
A generic reverse step from $\mathbf{x}_t$ to $\mathbf{x}_{t-1}$ consists of:
(1) \textbf{Predict:} run the model on $\mathbf{x}_t$ and $\mathbf{c}$ to obtain per-position token distributions;
(2) \textbf{Select:} choose an update set $\mathcal{U}_t \subseteq \mathcal{M}_t$ (which masked positions to fill or revise);
(3) \textbf{Commit:} sample or take $\arg\max$ on $\mathcal{U}_t$ to form $\mathbf{x}_{t-1}$, optionally allowing remasking or other corrections.
When $|\mathcal{U}_t|$ is large, many tokens are updated in parallel, which is the main pathway to potential speedups over left-to-right AR decoding.

Although the model may be evaluated on the full sequence, many practical schedules and policies change only a subset of positions per step.
Let $\Delta_t \subseteq \{1,\dots,L\}$ denote the positions whose token values actually change between steps $t$ and $t{-}1$ (often $\Delta_t\subseteq\mathcal{U}_t$).
When $|\Delta_t| \ll L$, the trajectory exhibits \emph{update sparsity}, a property exploited by both adaptive scheduling/policy choices and diffusion-aware reuse mechanisms surveyed later.
However, quantifying the actual speedups from these techniques requires disentangling multiple interacting factors such as step count, per-step compute, and systems overhead, which motivates the latency decomposition framework we introduce next.

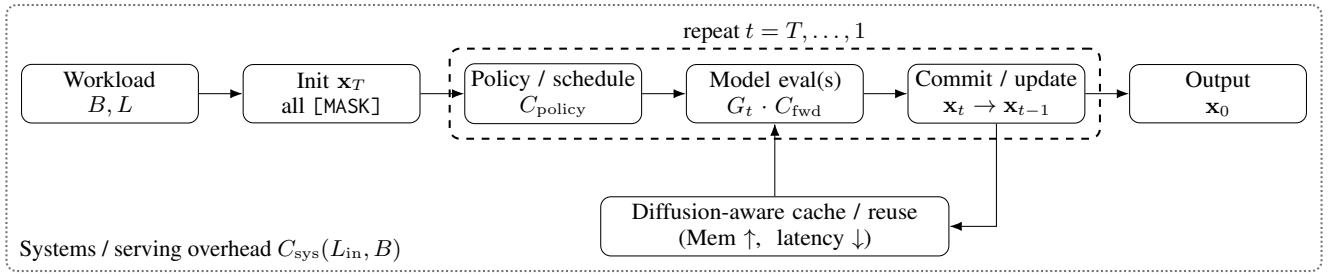
\begin{figure*}[t]
\centering
\small
\resizebox{0.98\textwidth}{!}{%
\begin{tikzpicture}[>=Latex, node distance=6mm]
\tikzset{
  block/.style={draw, rounded corners, align=center,
                minimum height=8mm, text width=2.25cm, inner sep=2.5pt},
  wide/.style={draw, rounded corners, align=center,
               minimum height=8mm, text width=5.8cm, inner sep=2.5pt},
  dashedbox/.style={draw, rounded corners, dashed, inner sep=2mm, line width=0.8pt},
  sysbox/.style={draw, rounded corners, densely dotted, inner sep=2mm,
                 draw=black!55, line width=0.8pt}
}

\node[block] (req) {Workload\\$B, L$};
\node[block, right=6mm of req] (init) {Init $\mathbf{x}_T$\\all \texttt{[MASK]}};
\node[block, right=6mm of init] (policy) {Policy / schedule\\$C_{\mathrm{policy}}$};
\node[block, right=6mm of policy] (model) {Model eval(s)\\$G_t\cdot C_{\mathrm{fwd}}$};
\node[block, right=6mm of model] (update) {Commit / update\\$\mathbf{x}_{t}\!\to\!\mathbf{x}_{t-1}$};
\node[block, right=6mm of update] (out) {Output\\$\mathbf{x}_0$};

\draw[->] (req) -- (init);
\draw[->] (init) -- (policy);
\draw[->] (policy) -- (model);
\draw[->] (model) -- (update);
\draw[->] (update) -- (out);

\node[dashedbox, fit=(policy) (model) (update)] (loop) {};
\node[font=\small, fill=white, inner sep=1pt] (replabel)
  at ([yshift=2mm]loop.north) {repeat $t=T,\dots,1$};

\node[wide, text width=4.6cm, below=10mm of model] (cache)
{Diffusion-aware cache / reuse\\(Mem $\uparrow$,\ \ latency $\downarrow$)};
\draw[->] (cache.north) -- (model.south);
\draw[->] (update.south) |- (cache.east);

\node[sysbox, fit=(req) (init) (loop) (out) (cache) (replabel)] (sys) {};
\node[font=\small, fill=white, inner sep=1pt, anchor=south west]
  at ([xshift=1.5mm,yshift=1mm]sys.south west)
  {Systems / serving overhead $C_{\mathrm{sys}}(L_{\mathrm{in}},B)$};

\end{tikzpicture}%
}
\caption{End-to-end inference schematic for masked/discrete diffusion decoding and the latency terms in Eq.~\ref{eq:latency_decomposition}. The reverse-time refinement loop (repeat $t=T,\dots,1$) alternates lightweight policy/schedule decisions ($C_{\mathrm{policy}}$) and $G_t$ model evaluations (each with cost $C_{\mathrm{fwd}}$, dependent on the processed length $L_{\mathrm{in}}$ and batch size $B$) to update $\mathbf{x}_t \!\to\! \mathbf{x}_{t-1}$. Diffusion-aware caching/reuse can reduce redundant per-step computation or orchestration at the cost of higher memory, while systems/serving overhead $C_{\mathrm{sys}}(L_{\mathrm{in}},B)$ applies to the entire request.}
\label{fig:latency_pipeline}
\end{figure*}

\section{Latency Decomposition and Measuring Inference Efficiency}
\label{sec:latency}

Efficiency claims for diffusion-based LLM inference are easy to misinterpret.
In masked dLLMs, the refinement step count $T$ is a coarse proxy. However, wall-clock latency depends on how many \emph{full-sequence} model evaluations are executed, how expensive each evaluation is under the chosen hardware/precision/length/batch regime, and how much overhead arises from scheduling policies and serving systems.
This section introduces a compact decomposition that we use as a unifying lens throughout the survey.

\paragraph{Latency decomposition.}
Consider conditional generation under prompt/context $\mathbf{c}$ with batch size $B$ and generated span length $L$.
Let $L_{\mathrm{in}}$ denote the total sequence length processed per model evaluation, typically prompt length plus the generated span.
Most masked/discrete decoders run an iterative loop for $T$ refinement steps, and each step typically performs at least one full-sequence model evaluation.
Moreover, inference-time mechanisms such as guidance or deliberate compute scaling can require multiple evaluations per nominal step \cite{Schiff2025Simple,Dang2025InferenceTime}.
We decompose end-to-end latency as:
\begin{equation}
\begin{split}
\mathrm{Latency}(L_{\mathrm{in}},B) &\approx \sum_{t=1}^{T} \Big( G_t \cdot C_{\mathrm{fwd}}(L_{\mathrm{in}},B) + C_{\mathrm{policy}}(t;L) \Big) \\
&\qquad \qquad + C_{\mathrm{sys}}(L_{\mathrm{in}},B),
\end{split}
\label{eq:latency_decomposition}
\end{equation}
where $C_{\mathrm{fwd}}(L_{\mathrm{in}},B)$ is the cost of one model forward pass; $G_t \ge 1$ is the number of forward passes executed at step $t$ (capturing multi-pass decoding); $C_{\mathrm{policy}}(t;L)$ captures schedule/policy overhead (e.g., update-set selection, remasking, stopping); and $C_{\mathrm{sys}}(L_{\mathrm{in}},B)$ captures systems overhead such as kernel dispatch, memory movement, cache management, and framework orchestration, which can be substantial in interactive and microbatching regimes \cite{Ma2025dInfer,Fan2025Taming}.
We also define the total number of model evaluations as
$N_{\mathrm{fwd}} \triangleq \sum_{t=1}^{T} G_t$.

Equation~\eqref{eq:latency_decomposition} clarifies that ``reducing $T$'' is only one lever.
Speedups can come from reducing $T$ (fewer refinement iterations), reducing $N_{\mathrm{fwd}}$ (less multi-pass compute), lowering $C_{\mathrm{fwd}}$ (architectural/numerical improvements), lowering $C_{\mathrm{policy}}$ (lighter scheduling and fewer synchronizations), or lowering $C_{\mathrm{sys}}$ (serving and orchestration optimizations).
Table~\ref{tab:efficiency_levers} summarizes how major technique families map to these terms, and Figure~\ref{fig:latency_pipeline} shows where each cost arises in the decoding pipeline.

Common pitfalls include treating $T$ as a speed proxy without accounting for $N_{\mathrm{fwd}}$, using ambiguous ``tokens/sec'' definitions, and timing without a consistent end-to-end protocol (warmup, synchronization, and clear inclusion/exclusion of host overhead).
In the remainder of the survey, we organize acceleration methods by which terms in Eq.~\eqref{eq:latency_decomposition} they primarily target, making trade-offs explicit.

\section{Algorithmic Efficiency: Reducing $T$ and $N_{\mathrm{fwd}}$}
\label{sec:algorithmic}

Algorithmic acceleration methods primarily reduce inference latency by decreasing the refinement iterations $T$ and/or the total number of model evaluations $N_{\mathrm{fwd}}=\sum_{t=1}^{T}G_t$ in Eq.~\ref{eq:latency_decomposition}.
They modify the \emph{sampling trajectory}---the schedule, the update policy, or the decoding algorithm---rather than the model kernels or serving stack.
In practice, realized gains depend not only on $T$ but also on policy overhead $C_{\mathrm{policy}}$ and synchronization costs (Section~\ref{sec:latency}).

\subsection{Advanced Schedules \& Policies}
\label{sec:algo_schedules}

Schedules control the corruption level across steps, while policies decide which positions to update and when to stop.
Both determine how much parallel progress each step makes, and thus how small $T$ can be at a fixed quality target.

\subsubsection{Non-uniform, dilated, and jump schedules}
Reducing $T$ can be achieved by reallocating steps non-uniformly along the reverse-time trajectory, e.g., skipping/dilating portions while maintaining stable refinement \cite{Luxembourg2025Plan,Amin2025Why}.
Related work studies faster discrete diffusion solvers or higher-order updates that approximate longer trajectories with fewer evaluations \cite{Chen2024Fast,Ren2025Fast}.
For meaningful comparisons, the mapping from ``nominal steps'' to actual evaluation count (including multi-stage updates) should be explicit, since it affects both $T$ and $N_{\mathrm{fwd}}$.

\subsubsection{Confidence/entropy-driven unmasking and early exit}
Adaptive policies use uncertainty signals (probability, entropy, margin) to choose update sets $\mathcal{U}_t$ and concentrate computation on difficult positions \cite{Mohamed2025FastDecoding,BenHamu2025Accelerated}.
Training-free variable-length denoising or early stopping further reduces the \emph{average} step count by terminating when confidence criteria are met \cite{Li2025Fixed}.
These methods can introduce nontrivial $C_{\mathrm{policy}}$ (scoring/sorting/thresholding) and implementation-sensitive overheads; thus, reporting should include $N_{\mathrm{fwd}}$ and step-count distributions (e.g., percentiles), not only mean $T$.

\subsubsection{Learned unmasking policies}
Learned policies predict which positions to update (or when to stop) via supervised learning or policy-gradient/RL objectives \cite{Jazbec2025Learning,Wang2025SPG}.
They can improve the quality--latency frontier, but should be reported with (i) the policy feature set, (ii) any extra model calls, and (iii) an ablation isolating policy overhead from gains due to fewer evaluations.

\subsection{Speculative \& Parallel Decoding}
\label{sec:algo_speculative}

Decoding-algorithm changes can reduce the \emph{effective} expensive computation per sequence by enabling larger jumps in sequence space or amortizing strong-model calls via verification.
In Eq.~\ref{eq:latency_decomposition}, these methods target $T\downarrow$ and/or reduce expensive evaluations, but the net gain depends on acceptance and orchestration cost.

\subsubsection{Draft-and-verify (speculative) diffusion decoding}
Speculative diffusion decoding produces candidate updates with a cheaper mechanism and verifies/corrects them with a stronger denoiser \cite{Christopher2025Speculative,Li2025DiffuSpec,Gao2025Self}.
Speedups depend on acceptance rate and the relative cost of draft vs verification, so compute accounting should separate draft and verify evaluations and report acceptance statistics alongside end-to-end latency.

\subsubsection{Block / set decoding and structured parallel updates}
Block/set decoding updates structured groups of tokens per iteration, effectively interpolating between AR and fully parallel refinement \cite{Gat2025Set,Arriola2025Block}.
These methods can reduce iterations at matched quality, but may increase per-step decision complexity; evaluation should clarify how the structure changes $\Delta_t$ and whether full-length $L_{\mathrm{in}}$ reprocessing is required each step.

\subsubsection{Hybrid diffusion--autoregressive decoding}
Hybrid decoders combine diffusion refinement with autoregressive components, e.g., diffusion for global proposals and AR for final emission \cite{Li2025ReFusion,Liu2025TiDAR}.
Because different modules may dominate runtime under different regimes, results should state which module dominates compute and how $L_{\mathrm{in}}$ differs across modules.

\subsection{Distillation \& Consistency}
\label{sec:algo_distill}

Distillation-based methods reduce inference cost by shifting work to training, aiming to shrink $T$ toward a small constant without increasing $G_t$.

\subsubsection{Step / sampler distillation}
Progressive and learnable sampler distillation compress multi-step sampling into fewer steps \cite{Salimans2022Progressive,Fu2025Learnable,Hayakawa2025Distillation}.
The main benefit is direct $T\downarrow$, while common caveats include sensitivity in calibration/controllability as trajectories become shorter.

\subsubsection{Consistency-style objectives}
Consistency objectives enable few-step (or direct) mapping from noisy templates to clean samples \cite{Song2023Consistency,Kim2025CDLM}.
They target the dominant $T\cdot C_{\mathrm{fwd}}$ term with minimal inference-time logic, but should be evaluated across step budgets to show graceful quality degradation as $T$ shrinks.

\subsubsection{Self-distillation through time and unrolled generation}
Self-distillation across steps and unrolled training encourage faster convergence in fewer iterations \cite{Savinov2022Stepunrolled,Deschenaux2025Autoregression}.
From an efficiency lens, the key is robustness across workloads and decoding configurations, not only the smallest achievable $T$.

\section{Architectural \& Systems Efficiency: Reducing $C_{\mathrm{fwd}}$ and $C_{\mathrm{sys}}$}
\label{sec:archsys}

Even with the same refinement iterations $T$ and evaluation count $N_{\mathrm{fwd}}$, end-to-end latency in Eq.~\ref{eq:latency_decomposition} can vary substantially across implementations and deployment regimes.
This section focuses on lowering the per-evaluation cost $C_{\mathrm{fwd}}(L_{\mathrm{in}},B)$ and the request-level overhead $C_{\mathrm{sys}}(L_{\mathrm{in}},B)$.
A key difference from autoregressive serving is that masked diffusion decoding repeatedly evaluates a (typically) full-sequence Transformer under gradually changing token patterns, which creates both opportunities (reuse across similar steps) and challenges (cache invalidation, orchestration overhead, memory pressure).

\subsection{Dynamic Compute \& Sparsity}
\label{sec:arch_sparsity}

Architectural and numerical optimizations primarily target $C_{\mathrm{fwd}}$ (and sometimes peak memory), aiming to make each refinement evaluation cheaper.
Unlike Section~\ref{sec:algorithmic}, these methods do not necessarily reduce $T$, but can yield large gains when $N_{\mathrm{fwd}}$ remains nontrivial.

\subsubsection{Sparse / structured computation for diffusion decoding}
Because diffusion decoding revisits similar intermediate states across steps, several works investigate sparsity patterns tailored to diffusion LMs, e.g., sparse attention mechanisms to reduce attention cost and memory footprint \cite{Wang2025SparseD}.
The practical benefit depends strongly on kernel maturity and how sparsity interacts with batching and sequence length; thus, wall-clock latency (not only FLOPs) should be reported under the intended runtime stack.

\subsubsection{Lightweight denoisers and decoding architectures}
Another route is to reduce $C_{\mathrm{fwd}}$ via architectural choices that maintain refinement behavior with lower per-step cost, such as convolutional decoding or alternative decoder structures for diffusion language modeling \cite{Seo2025Fast,Arriola2025EncoderDecoder}.
In practice, such designs should be evaluated together with the decoding configuration (steps/policy), since architectural changes can shift where the bottleneck lies (compute vs memory bandwidth vs orchestration).

\subsubsection{Numerics and quantization}
Quantization and reduced-precision execution can lower both compute and memory cost per evaluation, but require calibration that matches the diffusion decoding distribution across timesteps and masking patterns \cite{Xu2025DLLMQuant,Zhang2025QuantdLLM}.
For efficiency claims, it is important to report the precision mode, any accuracy recovery tricks, and whether the timed measurement includes dequantization/casting overheads inside the forward path.

\subsection{Diffusion-Aware Caching (KV \& Activation)}
\label{sec:arch_cache}

Caching and reuse aim to exploit redundancy across refinement steps, reducing effective $C_{\mathrm{fwd}}$ and sometimes $C_{\mathrm{sys}}$ at the cost of higher memory.
Unlike autoregressive decoding, diffusion decoding revises token values across the sequence; thus, naive KV caching can become stale and requires diffusion-specific refresh and eviction strategies.

\subsubsection{KV caching across refinement steps}
A growing line of work adapts KV caching to diffusion LMs by selectively reusing attention states across steps while accounting for token revisions \cite{Ma2025dKVCache,Wu2025FastdLLMa,Hu2025FlashDLM,Nguyen-Tri2025Attention}.
These approaches can reduce repeated attention computation when consecutive states are similar, but they introduce a clear memory--speed trade-off and can shift the bottleneck to memory bandwidth and cache management.

\subsubsection{Selective refresh and cache eviction}
Because only a subset of token values often changes per step (Section~\ref{sec:background}), caching can be made more effective by refreshing only where updates occur and evicting cache entries predicted to be unhelpful \cite{Huang2025Mask,Song2025SparsedLLM,Jiang2025d2Cache,Bu2025DiCache}.
A key evaluation point is whether caching remains beneficial under realistic serving regimes (microbatching, long-context prompts, and concurrent requests), where memory pressure and fragmentation can dominate.

\subsubsection{Reporting implications}
For diffusion-aware caching, a minimal report should include: (i) peak GPU memory with and without caching, (ii) wall-clock latency and throughput under matched $L_{\mathrm{in}}$ and concurrency, and (iii) quality at the exact caching configuration.
Because caching can also change $C_{\mathrm{sys}}$ (extra bookkeeping, synchronization, or memory movement), timing should follow an end-to-end protocol (Section~\ref{sec:latency}) rather than isolated kernel timings.

\subsection{System-Level Optimization}
\label{sec:arch_system}

System-level optimizations target $C_{\mathrm{sys}}(L_{\mathrm{in}},B)$ by reducing orchestration overheads and improving utilization across the repeated refinement loop.
These effects are often most visible for batch-1 latency and interactive settings, where dispatch and synchronization overheads can become comparable to compute.

\subsubsection{Diffusion-aware inference frameworks}
Dedicated inference frameworks for diffusion LMs aim to streamline the refinement loop, manage caching, and reduce per-step overheads that do not scale with model FLOPs \cite{Ma2025dInfer}.
From an efficiency perspective, the systems should be evaluated across batch sizes and concurrency levels to show when $C_{\mathrm{sys}}$ dominates versus when compute dominates.

\subsubsection{Production serving and memory dynamics}
Production-oriented work highlights that diffusion decoding can create distinct memory and orchestration challenges due to repeated full-sequence evaluations, cache growth, and step-wise control flow \cite{Fan2025Taming}.
Accordingly, system claims should report not only average latency but also tail latency (p95) and memory headroom under realistic request mixes.

\section{Inference-Time Scaling: The $G_t$ Factor}
\label{sec:scaling}

Many recent improvements in diffusion-based language generation come from \emph{inference-time scaling}: deliberately spending more computation at test time to improve quality, controllability, or constraint satisfaction.
In Eq.~\ref{eq:latency_decomposition}, these methods typically increase the per-step evaluation multiplier $G_t$ (hence $N_{\mathrm{fwd}}$), and often also increase $C_{\mathrm{policy}}$ and memory footprint due to maintaining additional trajectories, scores, or caches.
As a result, step counts alone can be especially misleading in this regime: two decoders with the same $T$ can differ substantially in wall-clock latency and peak memory depending on the scaling strategy.

\subsection{Guidance as Multi-Pass Decoding}
\label{sec:scaling_guidance}

Guidance mechanisms improve generation quality or enforce preferences by altering the reverse-time update using additional signals, often requiring multiple model evaluations per refinement step.
In masked/discrete diffusion, simple guidance variants can be implemented in a way that resembles classifier-free guidance (or related conditioning tricks), which naturally introduces a multi-pass structure \cite{Schiff2025Simple}.
Recent work also explores adaptive guidance rules that modulate the guidance strength or update pattern based on uncertainty, which can improve the quality--compute trade-off but makes $G_t$ and $C_{\mathrm{policy}}$ configuration-dependent \cite{Li2025Adaptive,Ye2025What}.
Remasking-based scaling can be viewed similarly: additional refinement passes (or repeated mask-and-refine cycles) increase effective compute beyond the nominal step count \cite{Wang2025Remasking}.

From an efficiency standpoint, guidance should be reported by explicitly stating: (i) how many forwards are executed per step (the resulting $G_t$ schedule), (ii) whether evaluations are performed on the full length $L_{\mathrm{in}}$, and (iii) the exact operating point (guidance strength / masking thresholds) at which quality is measured.
Without this, comparisons that only match $T$ can significantly under- or over-estimate runtime.

\subsection{Search and Trajectory-Level Scaling}
\label{sec:scaling_search}

A second scaling family treats diffusion decoding as an implicit search problem over refinement trajectories.
Instead of committing to a single reverse chain, these methods explore multiple candidates and select or resample trajectories based on scores, rewards, or constraints.
This often yields strong gains in alignment or constraint satisfaction, but increases compute roughly with the breadth/particles/expansions used, directly inflating $N_{\mathrm{fwd}}$.

\subsubsection{Particle / SMC-style scaling}
Particle-based methods (e.g., Particle Gibbs sampling) scale computation by maintaining and resampling a set of candidate trajectories, using the additional compute to better explore the posterior over discrete sequences \cite{Dang2025InferenceTime}.
In Eq.~\ref{eq:latency_decomposition}, the dominant effect is $G_t\uparrow$ (multiple evaluations per step and/or per particle), with additional policy overhead from resampling and scoring.

\subsubsection{Tree search and constrained inference}
Tree-search formulations explicitly branch on promising refinements and evaluate candidates with search policies, e.g., MCTS layered on top of diffusion decoding \cite{Huang2025Diffusion}.
Related work studies constrained decoding objectives for discrete diffusion, where the scaling budget is spent on ensuring constraints or optimizing utility under constraints \cite{Cardei2025Constrained,Suresh2025DINGO}.
Diffusion tree sampling provides another scalable framework for inference-time alignment by expanding candidate refinement paths \cite{Jain2025Diffusion}.
These approaches can be effective when constraints are hard to satisfy with a single-pass decoder, but their efficiency is highly regime-dependent: branching increases memory pressure (multiple partial states and caches) and can interact strongly with serving-level overheads (Section~\ref{sec:archsys}).

\section{Practitioner's Guide}
\label{sec:guide_and_challenges}

This section translates the surveyed techniques into a practical workflow grounded in Eq.~\eqref{eq:latency_decomposition}.
We do \emph{not} introduce a new benchmark nor report new measurements; instead, we summarize diffusion-specific knobs that determine inference-time compute and explain how to interpret efficiency claims beyond nominal step counts.

\paragraph{Core principle.}
In diffusion-based decoding, the refinement budget $T$ alone is not a reliable proxy for speed.
End-to-end latency depends on the total number of model evaluations $N_{\mathrm{fwd}}=\sum_{t=1}^{T} G_t$, the per-evaluation cost, and overhead from schedule/policy logic and iterative control flow (Eq.~\eqref{eq:latency_decomposition}).
Accordingly, when comparing methods, it is essential to make the effective compute budget explicit (via $T$ and $G_t$) and to compare at matched quality operating points.

\subsection{A Workflow Grounded in Eq.~\eqref{eq:latency_decomposition}}
\label{sec:guide_workflow}

We recommend the following workflow when selecting and evaluating acceleration methods.

\paragraph{Step 1: Fix the workload and objective.}
Specify prompt length and generated length (or their distributions), and define the processed length per model evaluation $L_{\mathrm{in}}$.
Clarify the objective: latency-sensitive generation, throughput-oriented serving, or quality/constraint-critical decoding.

\paragraph{Step 2: Make the compute budget explicit in diffusion terms.}
Report the refinement budget $T$ together with the per-step evaluation multiplier $G_t$ (hence $N_{\mathrm{fwd}}=\sum_t G_t$).
If the method introduces multi-pass evaluation (e.g., guidance, verification, particles, or branching), state how $G_t$ changes and what the effective evaluation count becomes.

\paragraph{Step 3: Choose levers that target the dominant term(s).}
Use Eq.~\eqref{eq:latency_decomposition} as a diagnostic lens:
methods in Section~\ref{sec:algorithmic} primarily reduce iterations ($T$) and/or evaluations ($N_{\mathrm{fwd}}$);
methods in Section~\ref{sec:archsys} reduce per-evaluation cost via numerics, structured compute, and reuse/caching across steps;
methods in Section~\ref{sec:scaling} often increase $G_t$ (thus $N_{\mathrm{fwd}}$) to improve quality or satisfy constraints.
In practice, the most meaningful comparison is to isolate which term(s) changed and what trade-offs were introduced.

\paragraph{Step 4: Compare on a small matched-quality frontier.}
When a method exposes a compute--quality knob (steps, early-exit thresholds, guidance strength, number of particles/branches),
reporting a few operating points makes the trade-off explicit and avoids comparisons at mismatched operating points.

\paragraph{Quick reference.}
Table~\ref{tab:decision_matrix} provides a compact map from common goals (latency, throughput, and quality-critical decoding) to the corresponding levers in Eq.~\eqref{eq:latency_decomposition} and the survey sections where representative methods are discussed.

\begin{table}[h]
\centering
\small
\renewcommand{\arraystretch}{1.15}
\setlength{\tabcolsep}{3pt}
\begin{tabularx}{\columnwidth}{l X l}
\toprule
\textbf{Bottleneck / Goal} & \textbf{Recommended Approach} & \textbf{Sec.} \\
\midrule
\rowcolor{AlgoBG}
\textbf{Latency-Sensitive} & \textbf{Minimize $T$ \& $C_{\mathrm{sys}}$} & \\
(Batch $\approx$ 1, Chat) & $\cdot$ Advanced Schedules & \S\ref{sec:algo_schedules} \\
& $\cdot$ System/Kernel Opt. & \S\ref{sec:arch_system} \\
\addlinespace[2pt]
\rowcolor{ArchBG}
\textbf{Throughput-Bound} & \textbf{Minimize $C_{\mathrm{fwd}}$ (Mem trade-off)} & \\
(Serving, Offline) & $\cdot$ Diffusion-aware Cache & \S\ref{sec:arch_cache} \\
& $\cdot$ Sparsity / Quantization & \S\ref{sec:arch_sparsity} \\
\addlinespace[2pt]
\rowcolor{ScaleBG}
\textbf{Quality-Critical} & \textbf{Scale $G_t$ (Multi-pass)} & \\
(Reasoning, Math) & $\cdot$ Guidance / Re-ranking & \S\ref{sec:scaling} \\
& $\cdot$ Verify w/ cheap Draft & \S\ref{sec:algo_speculative} \\
\bottomrule
\end{tabularx}
\caption{Decision matrix mapping deployment scenarios to primary efficiency factors in Eq.~\ref{eq:latency_decomposition}. We identify the dominant bottleneck for each case, including latency, throughput, and quality, and recommend corresponding acceleration families discussed in this survey.}
\label{tab:decision_matrix}
\end{table}

\subsection{Diffusion-Specific Notes for Interpreting Efficiency Claims}
\label{sec:guide_disclosure}

Most ambiguity in efficiency comparisons for diffusion decoding comes from under-specifying the refinement trajectory and the number of model evaluations.
The following diffusion-specific items are often sufficient to make comparisons interpretable without prescribing a particular measurement protocol:

\begin{itemize}
    \item \textbf{Trajectory and stopping:} $T$, the schedule type, and any early-exit criteria.
    \item \textbf{Evaluation accounting:} $G_t$ (or its description) and $N_{\mathrm{fwd}}=\sum_t G_t$, especially when guidance/search/verification is used.
    \item \textbf{Speculative or hybrid pipelines:} evaluation counts \emph{per module} (draft vs verify; diffusion vs AR) and acceptance/rejection statistics when applicable.
    \item \textbf{Update sparsity (recommended):} a compact statistic such as the average fraction of changed positions $|\Delta_t|/L$, since many reuse/caching mechanisms implicitly rely on sparsity across refinement steps (Section~\ref{sec:background}).
\end{itemize}

\subsection{Common Pitfalls for Interpreting Speedups}
\label{sec:guide_pitfalls}

Finally, we highlight recurring sources of ambiguity that can make ``speedup'' claims hard to interpret:

\begin{itemize}
    \item \textbf{$T$ reported without $G_t$:} identical step counts can correspond to very different compute budgets when multi-pass decoding is used.
    \item \textbf{Multi-pass logic hidden in a single iteration:} methods that pack multiple evaluations into one nominal step should make the effective $N_{\mathrm{fwd}}$ explicit.
    \item \textbf{Comparisons at mismatched operating points:} when a method trades compute for quality (or vice versa), reporting a small frontier is often more informative than a single tuned point.
\end{itemize}

Taken together, making $(T,\,G_t,\,N_{\mathrm{fwd}})$ explicit and tying claimed gains back to Eq.~\eqref{eq:latency_decomposition} is typically sufficient to keep diffusion inference efficiency comparisons robust and interpretable.

\section{Conclusion \& Open Challenges}
\label{sec:conclusion}

Diffusion-based large language models (dLLMs) offer a fundamentally different inference interface from autoregressive decoding: generation proceeds by iterative refinement with parallel updates over multiple token positions.
This creates a genuine opportunity for faster-than-AR decoding, but practical speedups require \emph{inference mechanisms} that turn parallel updates into fewer effective model evaluations, cheaper evaluations, and lower iteration overhead.

This survey focused on that inference-efficiency question.
We organized the acceleration landscape into three complementary axes.
First, \emph{algorithmic} techniques modify the refinement trajectory---schedules, update policies, decoding algorithms, and distillation---to reduce the iteration budget $T$ and/or the effective evaluation count $N_{\mathrm{fwd}}$ (Section~\ref{sec:algorithmic}).
Second, \emph{architectural and systems} techniques reduce the per-evaluation cost and amortize redundant computation across steps via numerics, structured compute, and diffusion-aware reuse (Section~\ref{sec:archsys}).
Third, \emph{inference-time scaling} methods deliberately increase compute (often via multi-pass evaluation) to improve quality or satisfy constraints, making the compute--quality trade-off explicit (Section~\ref{sec:scaling}).

A unifying lesson across these threads is that efficiency gains in dLLM inference are rarely attributable to a single knob.
Step reduction is most effective when it does not silently increase multi-pass computation ($G_t$), policy overhead, or sensitivity to decoding hyperparameters.
Reuse and caching can yield substantial wall-clock gains when consecutive states are similar, but they introduce explicit memory trade-offs and require diffusion-specific refresh/eviction logic.
Finally, scaling strategies can be powerful in quality-critical settings, but their benefits are meaningful only when the added evaluation budget is accounted for in the same currency as the base decoder (Eq.~\eqref{eq:latency_decomposition}).
To keep these trade-offs interpretable, we distilled a lightweight practitioner workflow and diffusion-specific disclosure items in Section~\ref{sec:guide_and_challenges}.

\textbf{Open challenges.}
We highlight several directions that, in our view, will most strongly shape the next phase of progress in accelerating diffusion-based language inference:

\begin{itemize}
    \item \textbf{Few-step decoding without brittle quality or control.}
    Distillation and consistency-style approaches can shrink $T$ dramatically, but robustness across prompts, lengths, and controllability settings remains uneven.
    An open problem is to achieve few-step generation that degrades gracefully with step budgets and remains well-calibrated under diverse decoding policies (Section~\ref{sec:algo_distill}).

    \item \textbf{Predictable acceleration from adaptive schedules and policies.}
    Adaptive unmasking and early-exit policies can reduce average compute, yet their benefits can be eroded by policy overhead or instability across workloads.
    Designing low-overhead policies with predictable speed--quality behavior (and clear failure modes) is central to making trajectory-level acceleration reliable (Section~\ref{sec:algo_schedules}).

    \item \textbf{Diffusion-aware reuse with correctness guarantees under token revisions.}
    Reuse mechanisms must remain valid when tokens are revised across steps.
    Better criteria for cache validity, selective refresh, and error control---especially under long contexts and memory constraints---are needed to make reuse both fast and dependable (Section~\ref{sec:arch_cache}).

    \item \textbf{Co-design of parallel updates and kernel-friendly execution.}
    Many decoders exhibit update sparsity ($|\Delta_t|\ll L$), yet most implementations still execute near full-sequence computation.
    Bridging diffusion-specific structure to hardware- and compiler-friendly execution (e.g., structured sparsity and efficient partial updates) is key to converting theoretical parallelism into consistent wall-clock gains (Section~\ref{sec:arch_sparsity}).

    \item \textbf{Compute-adaptive scaling for reasoning and constraints.}
    Guidance and search-based methods can improve quality, but they increase $G_t$ and can interact strongly with the base trajectory.
    A promising direction is compute-adaptive scaling that allocates additional evaluations only when needed, while keeping the compute budget and its effect on quality transparent (Section~\ref{sec:scaling}).
\end{itemize}

\paragraph{Limitations of this survey.}
This survey focuses on \emph{masked/discrete} diffusion language models, where the refinement loop involves explicit token-level updates. 
Continuous-embedding approaches (e.g., Diffusion-LM~\cite{Li2022DiffusionLM} and its variants) operate in a fundamentally different latent space and involve distinct efficiency trade-offs that warrant separate treatment.

Although this survey focuses on language generation, diffusion-style iterative refinement is also relevant to structured sequence domains such as time-series imputation and forecasting~\cite{Alcaraz2023Diffusion}. Recent work further applies large language diffusion models to time-series forecasting~\cite{Pei2025LEAF}.

Also, the field is evolving rapidly: many techniques surveyed here were published or released within the past year, and standardized benchmarks for dLLM inference efficiency remain lacking.
As a result, direct comparisons across papers are often confounded by differences in model scale, evaluation protocol, and hardware configuration.
While we have attempted to organize the literature through a unified framework (Eq.~\ref{eq:latency_decomposition}), we caution that reported speedups should be interpreted with attention to the specific experimental setup.

Finally, we restrict our attention to \emph{inference-time} efficiency. 
Training efficiency (\textit{e.g.,} data efficiency and adaptation from pretrained AR models) is an important but orthogonal concern that we do not address in depth.

\newpage


\section*{Acknowledgments}
This work was supported by the Institute for Information \& communications Technology Planning \& Evaluation (IITP) grants funded by the Korean government (MSIT) (RS-2019-II190075, Artificial Intelligence Graduate School Program (KAIST); RS-2025-02304967, AI Star Fellowship (KAIST); and RS-2024-00396828, Development of AI-based Low-Power 5G-A O-DU/O-CU; contribution rate: 33.3\%).

\bibliographystyle{named}
\bibliography{ijcai26}

@inproceedings{Amin2025Why,
  title = {Why {{Masking Diffusion Works}}: {{Condition}} on the {{Jump Schedule}} for {{Improved Discrete Diffusion}}},
  booktitle = {{{NeurIPS}}},
  author = {Amin, Alan N. and Gruver, Nate and Wilson, Andrew Gordon},
  year = 2025
}

@inproceedings{Arriola2025Block,
  title = {Block {{Diffusion}}: {{Interpolating Between Autoregressive}} and {{Diffusion Language Models}}},
  booktitle = {{{ICLR}}},
  author = {Arriola, Marianne and Gokaslan, Aaron and Chiu, Justin T. and others},
  year = 2025
}

@inproceedings{Arriola2025EncoderDecoder,
  title = {Encoder-{{Decoder Diffusion Language Models}} for {{Efficient Training}} and {{Inference}}},
  booktitle = {{{NeurIPS}}},
  author = {Arriola, Marianne and Schiff, Yair and Phung, Hao and Gokaslan, Aaron and Kuleshov, Volodymyr},
  year = 2025
}

@inproceedings{Austin2021Structured,
  title = {Structured {{Denoising Diffusion Models}} in {{Discrete State-Spaces}}},
  booktitle = {{{NeurIPS}}},
  author = {Austin, Jacob and Johnson, Daniel D. and Ho, Jonathan and Tarlow, Daniel and van den Berg, Rianne},
  year = 2021
}

@inproceedings{BenHamu2025Accelerated,
  title = {Accelerated {{Sampling}} from {{Masked Diffusion Models}} via {{Entropy Bounded Unmasking}}},
  booktitle = {{{NeurIPS}}},
  author = {{Ben-Hamu}, Heli and Gat, Itai and Severo, Daniel and Nolte, Niklas and Karrer, Brian},
  year = 2025
}

@misc{Bu2025DiCache,
  title = {{{DiCache}}: {{Let Diffusion Model Determine Its Own Cache}}},
  author = {Bu, Jiazi and Ling, Pengyang and Zhou, Yujie and others},
  year = 2025,
  note = {arXiv}
}

@inproceedings{Cardei2025Constrained,
  title = {Constrained {{Discrete Diffusion}}},
  booktitle = {{{NeurIPS}}},
  author = {Cardei, Michael and Christopher, Jacob K. and Hartvigsen, Thomas and Kailkhura and others},
  year = 2025
}

@inproceedings{Chen2024Fast,
  title = {Fast {{Sampling}} via {{Discrete Non-Markov Diffusion Models}} with {{Predetermined Transition Time}}},
  booktitle = {{{NeurIPS}}},
  author = {Chen, Zixiang and Yuan, Huizhuo and Li, Yongqian and Kou, Yiwen and Zhang, Junkai and Gu, Quanquan},
  year = 2024
}

@inproceedings{Christopher2025Speculative,
  title = {Speculative {{Diffusion Decoding}}: {{Accelerating Language Generation}} through {{Diffusion}}},
  booktitle = {{{NAACL}}},
  author = {Christopher, Jacob K. and Bartoldson, Brian R. and {Ben-Nun}, Tal and others},
  year = 2025
}

@misc{Dang2025InferenceTime,
  title = {Inference-{{Time Scaling}} of {{Diffusion Language Models}} with {{Particle Gibbs Sampling}}},
  author = {Dang, Meihua and Han, Jiaqi and Xu, Minkai and Xu, Kai and Srivastava, Akash and Ermon, Stefano},
  year = 2025,
  note = {arXiv}
}

@inproceedings{Deschenaux2025Autoregression,
  title = {Beyond {{Autoregression}}: {{Fast LLMs}} via {{Self-Distillation Through Time}}},
  booktitle = {{{ICLR}}},
  author = {Deschenaux, Justin and Gulcehre, Caglar},
  year = 2025
}

@misc{Fan2025Taming,
  title = {Taming the {{Memory Footprint Crisis}}: {{System Design}} for {{Production Diffusion LLM Serving}}},
  author = {Fan, Jiakun and Zhang, Yanglin and Li, Xiangchen and Nikolopoulos, Dimitrios S.},
  year = 2025,
  note = {arXiv}
}

@inproceedings{Fu2025Learnable,
  title = {Learnable {{Sampler Distillation}} for {{Discrete Diffusion Models}}},
  booktitle = {{{NeurIPS}}},
  author = {Fu, Feiyang and Guo, Tongxian and Liu, Zhaoqiang},
  year = 2025
}

@misc{Gao2025Self,
  title = {Self {{Speculative Decoding}} for {{Diffusion Large Language Models}}},
  author = {Gao, Yifeng and Ji, Ziang and Wang, Yuxuan and Qi, Biqing and Xu, Hanlin and Zhang, Linfeng},
  year = 2025,
  note = {arXiv}
}

@misc{Gat2025Set,
  title = {Set {{Block Decoding}} Is a {{Language Model Inference Accelerator}}},
  author = {Gat, Itai and {Ben-Hamu}, Heli and Havasi, Marton and Haziza, Daniel and Reizenstein, Jeremy and Synnaeve, Gabriel and {Lopez-Paz}, David and Karrer, Brian and Lipman, Yaron},
  year = 2025,
  note = {arXiv}
}

@inproceedings{Hayakawa2025Distillation,
  title = {Distillation of {{Discrete Diffusion}} through {{Dimensional Correlations}}},
  booktitle = {{{ICML}}},
  author = {Hayakawa, Satoshi and Takida, Yuhta and Imaizumi, Masaaki and Wakaki, Hiromi and Mitsufuji, Yuki},
  year = 2025
}

@misc{Hu2025FlashDLM,
  title = {{{FlashDLM}}: {{Accelerating Diffusion Language Model Inference}} via {{Efficient KV Caching}} and {{Guided Diffusion}}},
  author = {Hu, Zhanqiu and Meng, Jian and Akhauri, Yash and Abdelfattah, Mohamed S. and Seo, Jae-sun and Zhang, Zhiru and Gupta, Udit},
  year = 2025,
  note = {arXiv}
}

@misc{Huang2025Diffusion,
  title = {Diffusion {{Language Model Inference}} with {{Monte Carlo Tree Search}}},
  author = {Huang, Zheng and Ramnath, Kiran and others},
  year = 2025,
  note = {arXiv}
}

@misc{Huang2025Mask,
  title = {Mask {{Tokens}} as {{Prophet}}: {{Fine-Grained Cache Eviction}} for {{Efficient dLLM Inference}}},
  author = {Huang, Jianuo and Zhang, Yaojie and Yang, Yicun and Huang, Benhao and Qi, Biqing and Liu, Dongrui and Zhang, Linfeng},
  year = 2025,
  note = {arXiv}
}

@inproceedings{Jain2025Diffusion,
  title = {Diffusion {{Tree Sampling}}: {{Scalable}} Inference-Time Alignment of Diffusion Models},
  booktitle = {{{NeurIPS}}},
  author = {Jain, Vineet and Sareen, Kusha and Pedramfar, Mohammad and Ravanbakhsh, Siamak},
  year = 2025
}

@misc{Jazbec2025Learning,
  title = {Learning {{Unmasking Policies}} for {{Diffusion Language Models}}},
  author = {Jazbec, Metod and Olausson, Theo X. and others},
  year = 2025,
  note = {arXiv}
}

@misc{Jiang2025d2Cache,
  title = {d \textasciicircum 2{{ Cache}}: {{Accelerating Diffusion-Based LLMs}} via {{Dual Adaptive Caching}}},
  author = {Jiang, Yuchu and Cai, Yue and Luo, Xiangzhong and Fu, Jiale and Wang, Jiarui and Liu, Chonghan and Yang, Xu},
  year = 2025,
  note = {arXiv}
}

@misc{Kim2025CDLM,
  title = {{{CDLM}}: {{Consistency Diffusion Language Models For Faster Sampling}}},
  author = {Kim, Minseo and Xu, Chenfeng and Hooper, Coleman and Singh, Harman and Athiwaratkun, Ben and Zhang, Ce and Keutzer, Kurt and Gholami, Amir},
  year = 2025,
  note = {arXiv}
}

@misc{Labs2025Mercury,
  title = {Mercury: {{Ultra-Fast Language Models Based}} on {{Diffusion}}},
  author = {{Inception Labs} and Khanna, Samar and Kharbanda, Siddhant  and others},
  year = 2025,
  note = {arXiv}
}

@inproceedings{Li2022DiffusionLM,
  title = {Diffusion-{{LM Improves Controllable Text Generation}}},
  booktitle = {{{NeurIPS}}},
  author = {Li, Xiang Lisa and Thickstun, John and Gulrajani, Ishaan and Liang, Percy and Hashimoto, Tatsunori B.},
  year = 2022
}

@inproceedings{Li2025Adaptive,
  title = {Adaptive {{Classifier-Free Guidance}} via {{Dynamic Low-Confidence Masking}}},
  booktitle = {{{NeurIPS}}},
  author = {Li, Pengxiang and Yan, Shilin and others},
  year = 2025
}

@misc{Li2025DiffuSpec,
  title = {{{DiffuSpec}}: {{Unlocking Diffusion Language Models}} for {{Speculative Decoding}}},
  author = {Li, Guanghao and Fu, Zhihui and Fang, Min and Zhao, Qibin and Tang, Ming and Yuan, Chun and Wang, Jun},
  year = 2025,
  note = {arXiv}
}

@misc{Li2025Fixed,
  title = {Beyond {{Fixed}}: {{Training-Free Variable-Length Denoising}} for {{Diffusion Large Language Models}}},
  author = {Li, Jinsong and Dong, Xiaoyi and Zang, Yuhang and Cao, Yuhang and Wang, Jiaqi and Lin, Dahua},
  year = 2025,
  note = {arXiv}
}

@misc{Li2025ReFusion,
  title = {{{ReFusion}}: {{A Diffusion Large Language Model}} with {{Parallel Autoregressive Decoding}}},
  author = {Li, Jia-Nan and Guan, Jian and Wu, Wei and Li, Chongxuan},
  year = 2025,
  note = {arXiv}
}

@misc{Li2025Survey,
  title = {A {{Survey}} on {{Diffusion Language Models}}},
  author = {Li, Tianyi and Chen, Mingda and Guo, Bowei and Shen, Zhiqiang},
  year = 2025,
  note = {arXiv}
}

@misc{Liu2025TiDAR,
  title = {{{TiDAR}}: {{Think}} in {{Diffusion}}, {{Talk}} in {{Autoregression}}},
  author = {Liu, Jingyu and Dong, Xin and Ye, Zhifan and Mehta, Rishabh and Fu, Yonggan and Singh, Vartika and Kautz, Jan and Zhang, Ce and Molchanov, Pavlo},
  year = 2025,
  note = {arXiv}
}

@inproceedings{Lou2024Discrete,
  title = {Discrete {{Diffusion Modeling}} by {{Estimating}} the {{Ratios}} of the {{Data Distribution}}},
  booktitle = {{{ICML}}},
  author = {Lou, Aaron and Meng, Chenlin and Ermon, Stefano},
  year = 2024
}

@misc{Luxembourg2025Plan,
  title = {Plan for {{Speed}}: {{Dilated Scheduling}} for {{Masked Diffusion Language Models}}},
  author = {Luxembourg, Omer and Permuter, Haim and others},
  year = 2025,
  note = {arXiv}
}

@misc{Ma2025dInfer,
  title = {{{dInfer}}: {{An Efficient Inference Framework}} for {{Diffusion Language Models}}},
  author = {Ma, Yuxin and Du, Lun and Wei, Lanning and others},
  year = 2025,
  note = {arXiv}
}

@inproceedings{Ma2025dKVCache,
  title = {{{dKV-Cache}}: {{The Cache}} for {{Diffusion Language Models}}},
  booktitle = {{{NeurIPS}}},
  author = {Ma, Xinyin and Yu, Runpeng and Fang, Gongfan and Wang, Xinchao},
  year = 2025
}

@misc{Mohamed2025FastDecoding,
  title = {Fast-{{Decoding Diffusion Language Models}} via {{Progress-Aware Confidence Schedules}}},
  author = {Mohamed, Amr and Zhang, Yang and Vazirgiannis, Michalis and Shang, Guokan},
  year = 2025,
  note = {arXiv}
}

@misc{Nguyen-Tri2025Attention,
  title = {Attention {{Is All You Need}} for {{KV Cache}} in {{Diffusion LLMs}}},
  author = {{Nguyen-Tri}, Quan and Ranjan, Mukul and Shen, Zhiqiang},
  year = 2025,
  note = {arXiv}
}

@inproceedings{Nie2025Large,
  title = {Large {{Language Diffusion Models}}},
  booktitle = {{{NeurIPS}}},
  author = {Nie, Shen and Zhu, Fengqi and You, Zebin and others},
  year = 2025
}

@inproceedings{Ren2025Fast,
  title = {Fast {{Solvers}} for {{Discrete Diffusion Models}}: {{Theory}} and {{Applications}} of {{High-Order Algorithms}}},
  booktitle = {{{NeurIPS}}},
  author = {Ren, Yinuo and Chen, Haoxuan and others},
  year = 2025
}

@inproceedings{Sahoo2024Simple,
  title = {Simple and {{Effective Masked Diffusion Language Models}}},
  booktitle = {{{NeurIPS}}},
  author = {Sahoo, Subham Sekhar and Arriola, Marianne and others},
  year = 2024
}

@inproceedings{Salimans2022Progressive,
  title = {Progressive {{Distillation}} for {{Fast Sampling}} of {{Diffusion Models}}},
  booktitle = {{{ICLR}}},
  author = {Salimans, Tim and Ho, Jonathan},
  year = 2022
}

@inproceedings{Savinov2022Stepunrolled,
  title = {Step-Unrolled {{Denoising Autoencoders}} for {{Text Generation}}},
  booktitle = {{{ICLR}}},
  author = {Savinov, Nikolay and Chung, Junyoung and Binkowski, Mikolaj and Elsen, Erich and van den Oord, Aaron},
  year = 2022
}

@inproceedings{Schiff2025Simple,
  title = {Simple {{Guidance Mechanisms}} for {{Discrete Diffusion Models}}},
  booktitle = {{{ICLR}}},
  author = {Schiff, Yair and Sahoo, Subham Sekhar and others},
  year = 2025
}

@inproceedings{Seo2025Fast,
  title = {Fast and {{Fluent Diffusion Language Models}} via {{Convolutional Decoding}} and {{Rejective Fine-tuning}}},
  booktitle = {{{NeurIPS}}},
  author = {Seo, Yeongbin and Lee, Dongha and Kim, Jaehyung and Yeo, Jinyoung},
  year = 2025
}

@inproceedings{Song2023Consistency,
  title = {Consistency {{Models}}},
  booktitle = {{{ICML}}},
  author = {Song, Yang and Dhariwal, Prafulla and Chen, Mark and Sutskever, Ilya},
  year = 2023
}

@misc{Song2025SparsedLLM,
  title = {Sparse-{{dLLM}}: {{Accelerating Diffusion LLMs}} with {{Dynamic Cache Eviction}}},
  author = {Song, Yuerong and Liu, Xiaoran and Li, Ruixiao and Liu, Zhigeng and Huang, Zengfeng and Guo, Qipeng and He, Ziwei and Qiu, Xipeng},
  year = 2025,
  note = {arXiv}
}

@inproceedings{Suresh2025DINGO,
  title = {{{DINGO}}: {{Constrained Inference}} for {{Diffusion LLMs}}},
  booktitle = {{{NeurIPS}}},
  author = {Suresh, Tarun and Banerjee, Debangshu and Ugare, Shubham and Misailovic, Sasa and Singh, Gagandeep},
  year = 2025
}

@inproceedings{Wang2025Remasking,
  title = {Remasking {{Discrete Diffusion Models}} with {{Inference-Time Scaling}}},
  booktitle = {{{NeurIPS}}},
  author = {Wang, Guanghan and Schiff, Yair and Sahoo, Subham Sekhar and Kuleshov, Volodymyr},
  year = 2025
}

@misc{Wang2025SparseD,
  title = {{{SparseD}}: {{Sparse Attention}} for {{Diffusion Language Models}}},
  author = {Wang, Zeqing and Fang, Gongfan and Ma, Xinyin and Yang, Xingyi and Wang, Xinchao},
  year = 2025,
  note = {arXiv}
}

@misc{Wang2025SPG,
  title = {{{SPG}}: {{Sandwiched Policy Gradient}} for {{Masked Diffusion Language Models}}},
  author = {Wang, Chenyu and Rashidinejad, Paria and Su, DiJia and others},
  year = 2025,
  note = {arXiv:2510.09541}
}

@misc{Wu2025FastdLLMa,
  title = {Fast-{{dLLM}}: {{Training-free Acceleration}} of {{Diffusion LLM}} by {{Enabling KV Cache}} and {{Parallel Decoding}}},
  author = {Wu, Chengyue and Zhang, Hao and others},
  year = 2025,
  note = {arXiv}
}

@misc{Xu2025DLLMQuant,
  title = {{{DLLMQuant}}: {{Quantizing Diffusion-based Large Language Models}}},
  author = {Xu, Chen and Yang, Dawei},
  year = 2025,
  note = {arXiv}
}

@misc{Ye2025Dream,
  title = {Dream {{7B}}: {{Diffusion Large Language Models}}},
  author = {Ye, Jiacheng and Xie, Zhihui and Zheng, Lin and Gao, Jiahui and Wu, Zirui and Jiang, Xin and Li, Zhenguo and Kong, Lingpeng},
  year = 2025,
  note = {arXiv}
}

@misc{Ye2025What,
  title = {What {{Exactly Does Guidance Do}} in {{Masked Discrete Diffusion Models}}},
  author = {Ye, He and Kevin, Rojas and Molei, Tao},
  year = 2025,
  note = {arXiv:2506.10971}
}

@misc{Zhang2025QuantdLLM,
  title = {Quant-{{dLLM}}: {{Post-Training Extreme Low-Bit Quantization}} for {{Diffusion Large Language Models}}},
  author = {Zhang, Tianao and Li, Zhiteng and Yan, Xianglong and Qin, Haotong and Guo, Yong and Zhang, Yulun},
  year = 2025,
  note = {arXiv}
}

@misc{Zhang2025Survey,
  title = {A {{Survey}} on {{Parallel Text Generation}}: {{From Parallel Decoding}} to {{Diffusion Language Models}}},
  author = {Zhang, Lingzhe and Fang, Liancheng and others},
  year = 2025,
  note = {arXiv}
}

@article{Alcaraz2023Diffusion,
  title = {Diffusion-based Time Series Imputation and Forecasting with Structured State Space Models},
  author = {Alcaraz, Juan Miguel Lopez and Strodthoff, Nils},
  journal = {{{TMLR}}},
  year = 2023
}

@inproceedings{Pei2025LEAF,
  title = {{LEAF}: Large Language Diffusion Model for Time Series Forecasting},
  author = {Pei, Yuhang and Ren, Tao and Wang, Yifan and Sun, Zhipeng and Ju, Wei and Chen, Chong and Hua, Xian-Sheng and Luo, Xiao},
  booktitle = {{{EMNLP}}},
  year = 2025
}

\end{document}